\pdfoutput=1

\documentclass[11pt]{article}
\usepackage{natbib}
\setcitestyle{authoryear, round, maxbibnames=2}

\usepackage[final]{acl}

\usepackage{times}
\usepackage{latexsym}
\usepackage{newtxtext,newtxmath}

\usepackage[T1]{fontenc}

\usepackage[utf8]{inputenc}

\usepackage{microtype}

\usepackage{inconsolata}

\usepackage{graphicx}
 \usepackage{amsmath}
 \usepackage{algorithm}
\usepackage{algorithmic}
\usepackage{booktabs}

\usepackage{pgfplots}
\pgfplotsset{compat=newest}
\usepackage{pgfplotstable}
\usepackage{caption}
\usepackage{subcaption}
\usepackage{soul}

\usepackage{colortbl}
\usepackage{xcolor}

\usepackage{xcolor,colortbl}
\usepackage{tikz}

\usepackage{longtable}
\usepackage{geometry}
\usepackage{lipsum}
\usepackage{multicol}
\usepackage{graphicx}
\usepackage{caption}
\usepackage{array}
\usepackage{lmodern}
\usepackage{booktabs}

\usepackage{pgfplots}
\usepackage{tikz}
\usepackage{subcaption}
\usepackage{adjustbox}
\usetikzlibrary{patterns}

\usepackage{tcolorbox}
\usepackage{graphicx}
\usepackage{multirow}
\usepackage{array}  

\title{Benchmarking Contextual and Paralinguistic Reasoning in Speech-LLMs: A Case Study with In-the-Wild Data}

\author{
  \textbf{Qiongqiong Wang},
  \textbf{Hardik B. Sailor},
  \textbf{Tianchi Liu},
  \textbf{Wenyu Zhang},
\\
  \textbf{Muhammad Huzaifah},
  \textbf{Nattadaporn Lertcheva},
  \textbf{Shuo Sun},
\\
  \textbf{Nancy F. Chen},
  \textbf{Jinyang Wu},
  \textbf{AiTi Aw}
  \\
\\
  Institute of Infocomm Research (I\textsuperscript{2}R), A\textsuperscript{$\star$}STAR, Singapore
 \\
  \small{
    \href{mailto:email@domain}{wang\underline{\enskip}qiongqiong@i2r.a-star.edu.sg}
  }
}

\begin{document}
\maketitle
\begin{abstract}
Recent speech-LLMs have shown impressive performance in tasks like transcription and translation, yet they remain limited in understanding the paralinguistic aspects of speech crucial for social and emotional intelligence. We propose CP-Bench, a benchmark for evaluating speech-LLMs on contextual paralinguistic reasoning the integration of verbal content with non-verbal cues like emotion and prosody. The benchmark includes two curated question-answering (QA) datasets requiring both linguistic and empathetic understanding. We evaluate state-of-the-art speech-LLMs from both open- and closed-source models and perform a comprehensive analysis across different question types. The top two models were further analyzed under temperature tuning to understand its effect on this task. Our benchmark reveals a key gap in existing evaluations and offers insights into building more context-aware and emotionally intelligent speech-capable LLMs.
\end{abstract}

\section{Introduction}
Large Language Models (LLMs) have achieved remarkable success in a wide range of natural language processing tasks, and recent advances have extended their capabilities into the multimodal domain, including vision, audio, and speech~\cite{Huang2024}. In particular, the integration of speech encoders with text-based LLMs, referred to as speech-LLMs, has enabled models to process and reason over spoken language input~\cite{Peng2024}. Recent developments in speech-LLMs demonstrate their ability to capture and generate semantically rich textual output from spoken input, achieving state-of-the-art performance in tasks such as automatic speech recognition, speech translation, and spoken question answering. 

Despite this success in modeling linguistic content, current speech-LLMs largely overlook the paralinguistic dimensions of human speech, such as emotion, intent, speaker identity, and social dynamics, which are essential for fully understanding and interpreting spoken communication. These paralinguistic cues are crucial for applications involving empathy, social reasoning, and human behavior analysis, yet they remain underrepresented in current evaluation benchmarks and model objectives. Driven by this potential, researchers have begun developing empathetic speech-LLMs that not only transcribe or interpret speech but also understand the speaker’s affective state and contextual nuances. Several studies have attempted to train models to understand emotions in speech and respond empathetically~\cite{Lin24,wang24blsp, kimparalinguistics, kang2024frozen}. Efforts have also extended to data creation; for example, a contextual paralinguistic data generation pipeline has been proposed that leverages LLMs to automatically produce QA pairs~\cite{cpqa_interspeech,cpqa_asru}.

Existing benchmarks for evaluating speech-LLMs primarily target isolated contextual tasks like spoken question answering (SQA) or paralinguistic tasks such as emotion recognition and speaker identification~\cite{audiobench,AIRbench,dynamicsuperb,OpenASQA,MMAU}. While valuable, these benchmarks fall short in assessing contextual paralinguistic reasoning, which is the integration of speech content, emotional nuance, and speaker-specific traits, which necessitates a more complete understanding of the input.

This gap motivates the need for a new benchmark that moves beyond isolated capabilities and evaluates the holistic reasoning ability of speech-LLMs in multimodal contexts. We introduce the task of contextual paralinguistic question answering (CPQA), which requires reasoning over both linguistic content and paralinguistic cues such as prosody, emotion, and speaker intent. For example, consider the utterance \textit{``There are so many people here. I used to come with my sister every summer.''} Without paralinguistic cues, a question like ``Why was the speaker happy?'' is ambiguous. The first phrase might signal excitement or frustration depending on tone, while the latter could convey nostalgia and either sadness or joy.  Resolving such ambiguity requires interpreting not only the transcript but also the speaker's tone and affective state. This highlights the necessity of prosody-dependent cues for capturing nuances beyond text. Such questions are also distinct from direct queries like ``What is the speaker’s emotion?'' as they demand deeper contextual reasoning about causes and states. In short, the evaluations demand a nuanced understanding of both what is said and how it is said. To this end, we propose \textit{CP-Bench}, a novel benchmark specifically designed to assess these capabilities in speech-LLMs. Our benchmark comprises two curated QA datasets derived from long and short in-the-wild speech segments. 


In addition to constructing and evaluating several state-of-the-art speech-LLMs on this proposed benchmark, we analyzed the impact of the temperature hyperparameter. Temperature controls the randomness of the generated response and significantly influences reasoning quality.
While higher temperatures can lead to diverse but potentially inconsistent responses, lower temperatures generally produce more factual and stable outputs. This stability is particularly crucial for tasks involving emotion or gender inference~\cite{renze-2024-effect}. To better understand this trade-off, we systematically analyzed how temperature settings affect model performance across different categories of contextual and paralinguistic question answering. Our goal is to identify optimal temperature ranges that balance output diversity with reasoning reliability, thereby enhancing the interpretability and robustness of speech-LLM responses in multimodal evaluation scenarios.

To our knowledge, this work presents the first comprehensive benchmark explicitly targeting reasoning over both context and paralinguistic cues in speech. Our main contributions are as follows:
\begin{itemize}
    \item We propose and construct CP-Bench, a benchmark designed to evaluate contextual-paralinguistic reasoning capabilities in speech-LLM models, using two question-answering (QA) sets derived from long and short speech segments. The benchmarking dataset will be shared with the wider community to spur more holistic benchmarking efforts\footnote{\url{https://huggingface.co/datasets/MERaLiON/CPQA-Evaluation-Set}}.
    
    \item We provide a thorough comparison and analysis of state-of-the-art open- and closed-sourced speech-LLMs, together with baseline speech-to-text systems consisting of cascading a speech encoder with an LLM, on the contextual paralinguistic question answering task.
    \item We investigate the effect of LLM temperature settings across different question types to understand their impact on model behavior.
\end{itemize}
Our findings reveal critical limitations in current models and offer insights for developing more intelligent, empathetic, and human-aligned speech-LLMs. CP-Bench sets a new direction for multimodal reasoning research at the intersection of speech and language understanding.

\section{Related Work}
\subsection{Multimodal Speech-LLMs}
Recent speech-LLMs, such as GPT-4~\cite{gpt4}, Qwen-audio~\cite{qwen,qwen2}, SALMONN~\cite{salmonn}, and MERaLiON-AudioLLM~\cite{meralion,cpqa_asru}, have demonstrated remarkable performance in handling of speech-based tasks. Some speech-LLMs, in particular, focus on contextual reasoning properties derived from speech~\cite{wang24blsp, Rubenstein23,Chen23,Lin24, Wang2023BLSPBL}. 
    
Several studies have attempted to train models to understand emotions in speech and respond empathetically~\cite{Lin24,wang24blsp, kimparalinguistics, kang2024frozen}. Among these, notable approaches introduced training strategies to enhance QA performance by incorporating paralinguistic information from existing speech emotion datasets~\cite{wang24blsp,kang2024frozen}. These models, however, exhibit limited capabilities in contextual reasoning alongside paralinguistic understanding, primarily because existing QA datasets lack coverage of both aspects. To incorporate paralinguistic cues, QA generation must extend beyond linguistic features. We refer to such QA as contextual paralinguistic QA (CPQA). 
    
\subsection{Audio/Speech Benchmarks}
Existing benchmarks such as AudioBench~\cite{audiobench}, Dynamic-Superb~\cite{dynamicsuperb}, AIR-Bench~\cite{AIRbench}, OpenASQA~\cite{OpenASQA}, and MMAU~\cite{MMAU} have made notable strides in evaluating speech-LLMs across various tasks, including transcription, speaker identification, and emotion recognition. Many of these benchmarks include question-answering components derived from well-known speech emotion datasets such as IEMOCAP~\cite{iemocap} and MELD~\cite{meld}. However, these QA tasks are typically framed in a direct and isolated manner, such as asking for an explicit emotion label, without requiring the model to perform higher-level contextual or inferential reasoning.

\section{The Proposed CP-Bench}
We propose a contextual-paralinguistic question answering benchmark (CP-Bench) to evaluate contextual-paralinguistic reasoning capabilities in speech-LLM models, using two QA sets derived from long and short speech segments. 


\subsection{Question Categories and Types}
\label{sssec:q-type}
We categorize question types into three primary categories based on their informational focus: content-only (C), direct paralinguistic (DP), and contextual paralinguistic (CP): 
\begin{itemize}
    \item \textbf{C} questions assess understanding of linguistic content alone, without relying on any paralinguistic cues.
    \item \textbf{DP} questions directly query paralinguistic attributes—such as emotion, speaker gender, or speaker count -- based solely on the auditory signal, without requiring contextual interpretation.
    \item \textbf{CP} questions require the integration of both linguistic content and paralinguistic information to arrive at an appropriate answer.
\end{itemize}
While speech-LLMs have demonstrated strong performance on C-type questions, most prior work addressing paralinguistics has primarily focused on the DP category. In contrast, the CP question category remains largely underexplored. Our work specifically targets this gap by evaluating models on CP-category questions.

To enable a more fine-grained analysis of paralinguistic understanding in speech QA, we 
further decompose the DP and CP categories into subtypes, which capture the specific 
attribute under inquiry: emotion (DE/CE), gender (DG/CG), and speaker count (DS/CS). 
Definitions for the question categories and associated types are summarized in Table~\ref{tab:qa_question_types}.

\begin{table*}[!t]
\centering
\resizebox{\textwidth}{!}{%
\begin{tabular}{ccl}
\toprule
\textbf{Category} & \textbf{Type} & \textbf{Definition} \\
\midrule
C & C & Questions about the contents of the speech or reasoning based on its contents. \\
\midrule
\multirow{3}{*}{DP} 
& DE & Questions that ask directly about the emotion in the speech. \\
& DS & Questions about the number of speakers in the conversation. \\
& DG & Questions that ask directly about the gender of the speaker(s). \\
\midrule
\multirow{3}{*}{CP} 
& CE & Requires finding emotion-related content and reasoning from it. \\
& CG & Requires gender-related reasoning from speech content. \\
& CS & Requires separating the speech by speaker and reasoning from content. \\
\bottomrule
\end{tabular}
}
\caption{Definitions of question categories and types used in the speech QA datasets.}
\label{tab:qa_question_types}
\end{table*}


\subsection{Evaluation Dataset Creation}
To ensure coverage of all targeted question types, we prioritized collecting emotionally rich speech, as such content is underrepresented in existing resources.  Emotionally salient clips were identified using data condensation methods~\cite{cpqa_interspeech}. Accordingly, two sets of 480 speech samples each were initially selected to form a long set (30-60 seconds) and short set (10-30 seconds) in order to evaluate model behavior across differing durations. 

Emotion and gender labels were derived through a two-step process: (1) automatic estimation, followed by (2) human correction and verification. For automatic estimation, we employed the Emotion2Vec speech emotion recognition (SER) tool\footnote{\url{https://github.com/ddlBoJack/emotion2vec}}~\cite{emotion2vec} for emotion labeling and a fine-tuned WavLM-ECAPA model for gender labeling following ~\cite{cpqa_interspeech}. The emotion taxonomy included angry, disgusted, fearful, happy, neutral, sad, and surprised. Gender labels were male, female, or unknown, with the latter assigned in low-confidence cases. Three human annotators then reviewed each audio clip. They corrected the labels if needed and were provided with detailed definitions of emotions in speech to ensure consistency during human annotation. The same human annotators subsequently generated QA pairs for each clip, following a unified guideline (see \ref{ssec:anno_guideline}). This protocol helped ensure consistency and reduced individual bias. 

\begin{table}
  \centering
  \small
  \begin{tabular}{@{}lcc@{}}
    \toprule
    \textbf{Type} & \textbf{Long} & \textbf{Short} \\
    \midrule
    C         & 124   & 70  \\
    DE        & 479   & 468    \\
    DG        & 66    & 89 \\
    DS      & 127   & 150 \\
    CE        & 574   & 302 \\
    CG        & 127   & 88 \\
    CS  & 70    & 13    \\
    \midrule
    \textbf{Total} & 1,567	& 1,180  \\
    \bottomrule
  \end{tabular}
  \caption{Statistics of question types.}
  \label{tab:qa_type}
\end{table}

For each speech sample, annotators generated at least two QA pairs. The QA pairs require reasoning over both paralinguistic (emotion, gender, sentiment, speaker traits) and verbal content. Annotators were also provided with transcriptions from the WhisperX speech recognition tool~\cite{Bain2023WhisperXTS} to facilitate processing. Each QA was tagged with one of the seven types listed in Table~\ref{tab:qa_question_types} for category-wise analysis. Annotators also ensured linguistic correctness, aided by ChatGPT for optional refinement. Furthermore, one emotion-query question was generated  per speech sample using a template randomly selected from those shown in Figure~\ref{fig:de_qa} in Appendix~\ref{ssec:appendix_prompts}. Following the above protocol, and after discarding the speech samples that lacked sufficient context for the required questions, we curated 1,567 QA pairs from 479 speech samples for the long set and 1,180 QA pairs from 468 samples for the short set. A statistical breakdown of the two datasets in terms of question type is shown in Table~\ref{tab:qa_type}. The number of CS-type QA pairs in the short set is small primarily because such brief audio segments are less likely to contain multiple speakers, leading to fewer relevant questions for annotation.

\begin{table*}[!t]
\centering
\small
\begin{tabular}{@{}l|r|cccccc@{}}
\toprule

\textbf{Set} & \textbf{Type} & \textbf{GPT} & \textbf{Gemini} & \textbf{Qwen} & \textbf{MERaLiON} & \textbf{Cascade 1} & \textbf{Cascade 2}\\
\midrule
 & C & \textbf{69.68} & 61.29 & 55.45 & \underline{62.10} & 54.39 & 61.81 \\
\textbf{Long}& CP & \textbf{67.48} & \underline{61.80} & 50.89 & 59.93 & 51.49 & 56.23\\
 & DP & 30.34 & 60.28 & \textbf{62.29} & \underline{60.80} & 33.30 & 44.18 \\
\midrule
 & C & \underline{72.57} & 66.69 & 56.39 & \textbf{73.69} & 45.94 & 40.74 \\
 \textbf{Short} & CP & \textbf{68.74} & 64.24 & 48.65 & \underline{68.30} & 49.23 & 64.43 \\
 & DP & 29.33 & \underline{66.27} & 52.39 & \textbf{67.95} & 35.88 & 45.58 \\

\bottomrule
\end{tabular}
\caption{Performance of all models using their default temperature settings on the long and short evaluation sets. The bold and underlined values indicate the best and second-best scores, respectively. }
\label{tab:models_comparison_group}
\end{table*}

\subsection{Evaluation Metrics}
\label{sssec:matric}
To evaluate model performance across all question types, we assess answer correctness using an LLM-based judge. A specialized prompt, tailored for contextual-paralinguistic QA, guides this evaluation. The prompt incorporates both linguistic content and paralinguistic cues (e.g., emotion, gender) and is carefully designed to avoid penalizing brief but accurate responses. Details are provided in Figure~\ref{fig:prompt_qa_dim}.

To evaluate overall performance across multiple sub-sets (e.g., question types), we adopt an Unweighted Average Score (UAS)
\begin{equation}
\text{UAS} = \frac{1}{K} \sum_{k=1}^{K} \left( \frac{1}{N_k} \sum_{n=1}^{N_k} S_{k,n} \right)
\end{equation}
where $K$ is the number of subsets (e.g., question types), $N_k$ is the number of QA pairs in sub-set $k$.  $S_{k,n}$ is the score for the $n$-th QA pair in subset $k$. This metric ensures a fair comparison between sets, particularly because the number of QA pairs varies across question types in both the long and short sets. UAS treats each subset equally, regardless of its size, thereby preventing any single type from disproportionately influencing the overall metric. This approach is analogous to unweighted accuracy in classification tasks, where each class is given equal importance regardless of its frequency in the dataset.

\section{Experiments}
We evaluate state-of-the-art speech-LLMs on the proposed CP-Bench, focusing on their performance across various question types. Additionally, we examine how temperature settings influence model behavior, using two of the top-performing models.

\subsection{Experimental Setup}
 Our evaluation includes four representative speech-LLMs, spanning closed-source, open-source, and cascaded approaches:
\begin{itemize}
    \item Closed-source: GPT-4o (audio-preview, API version: 2025-01-01-preview from Azure)\footnote{\url{https://learn.microsoft.com/en-us/azure/ai-services/openai/}} 
    and Gemini Flash 2.0\footnote{\url{https://ai.google.dev/gemini-api/docs/models\#gemini-2.0-flash}}~\cite{gemini}.  
    \item Open-source models: MERaLiON-AudioLLM-Whisper-SEA-LION\footnote{\url{https://huggingface.co/MERaLiON/MERaLiON-AudioLLM-Whisper-SEA-LION}}~\cite{meralion} 
    and Qwen2-Audio-7B-Instruct\footnote{\url{https://huggingface.co/Qwen/Qwen2-Audio-7B-Instruct}}~\cite{qwen2}.
    \item Cascade setup: Whisper large-v3\footnote{\url{https://huggingface.co/openai/whisper-large-v3}}~\cite{radford2022whisper} cascaded with (1) Gemma2 9B Instruct\footnote{\url{https://huggingface.co/google/gemma-2-9b-it}}~\cite{gemma_2024} (Cascade 1) and (2) Gemma2 9B CPT Sea Lion V3 Instruct\footnote{\url{https://huggingface.co/aisingapore/Gemma-SEA-LION-v3-9B-IT}} (Cascade 2).
\end{itemize}
We refer to these models as GPT, Gemini, MERaLiON, Qwen, Cascade 1, and Cascade 2 throughout the paper.

Model outputs are evaluated using LLaMA-3-70B-Instruct-AWQ\footnote{\url{https://huggingface.co/casperhansen/llama-3-70b-instruct-awq}}~\cite{llama3} as a judge. The maximum context length for the loaded model is set to 4,096 tokens, encompassing both the input prompt and the generated output. The maximum number of new tokens generated by the judge per response is capped at 512. For generation, temperature and top\_p were set to their default values of $1.0$ and $1.0$, respectively. Each response is scored on a scale from 0 to 5 using the evaluation prompt described in Section~\ref{sssec:matric}, and the scores are linearly rescaled to a 0--100 range for interpretability. To reduce variance, results are averaged over five independent runs of the judging process for each model's predicted set, ensuring the reliability of our reported average scores. Due to Qwen2's 30-second input limit, we evaluate both the first and last 30 seconds of each audio clip for the long set, reporting the higher score for each QA pair. Evaluation and judging were conducted on a single NVIDIA H100 80GB GPU for all models.

\subsection{Results and Analysis}
\subsubsection{Investigation across Models on Question Categories}
\begin{figure*}[!t]
\centering
\includegraphics[width=1\textwidth]{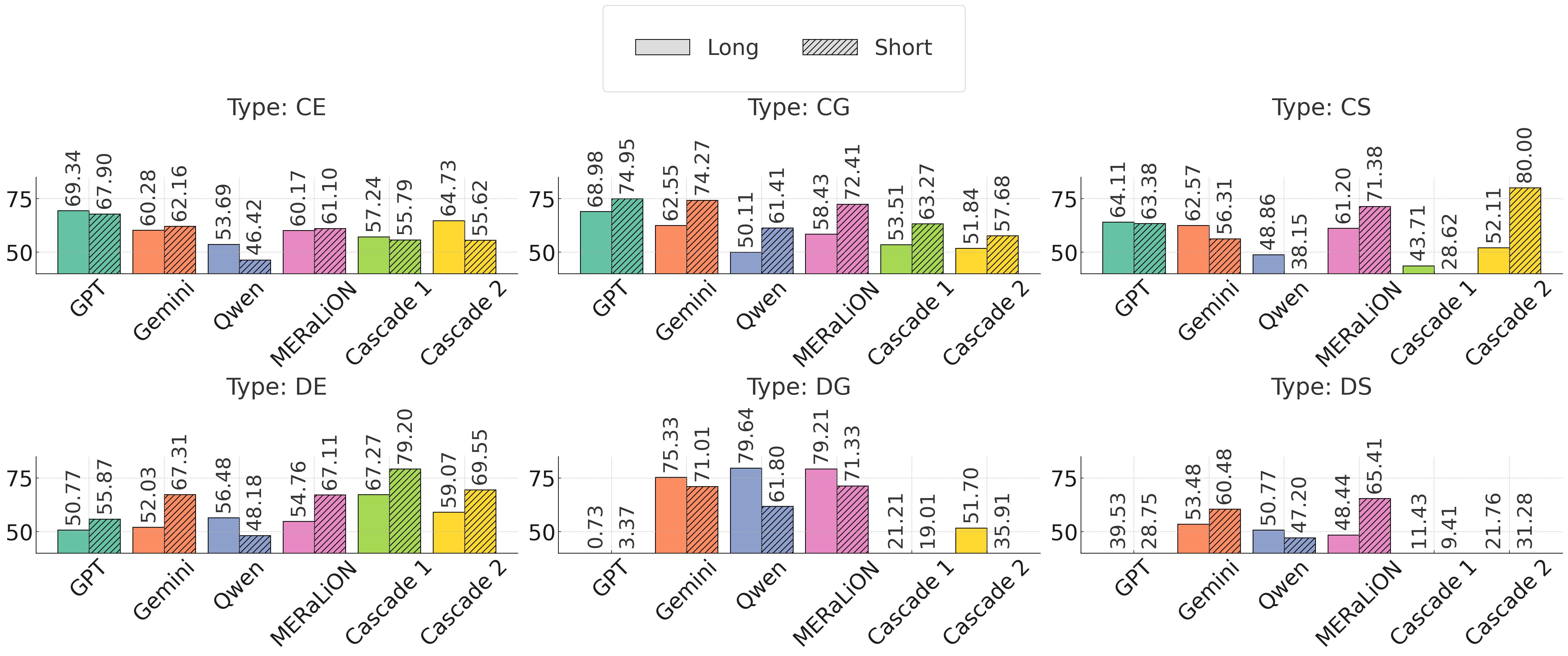}
\caption{Performance of all models using their default temperature settings on the long and short evaluation sets. For better visual comparison of top-performing models, the y-axis is limited to the 40–80 range. Scores below 40 are not shown as bars but are annotated numerically. }
\label{fig:all_model}
\end{figure*}

We first compare model performance on the high-level question categories using each model’s default temperature settings: 1.0 for all except 0.7 for GPT.  Table~\ref{tab:models_comparison_group} summarizes the UAS results across the three question categories for both the long and short evaluation sets. Some examples of QA pairs from the long set, generated by GPT and MERaLiON, are shown in Tables \ref{tab:qa_examples} and \ref{tab:qa_examples_meralion} in the Appendix.

%
GPT achieves the highest performance on contextual-only (C) and contextual-paralinguistic (CP) questions in both the long and short sets. However, it significantly underperforms  on direct paralinguistic (DP) questions, which indicates a limitation in directly recognizing paralinguistic cues such as emotion. The Microsoft Azure OpenAI API service includes content filtering and was found to be unable to answer questions related to speaker characteristics\footnote{\url{https://learn.microsoft.com/en-us/azure/ai-services/openai/concepts/content-filter}}. 

Gemini and MERaLiON demonstrate highly competitive and balanced performance across all categories  in the long set. In the short set, MERaLiON not only outperforms Gemini but also slightly surpasses GPT on C-category questions and performs comparably on CP-category questions. This suggests MERaLiON's strength in handling short-form speech input.
\begin{figure*}[t]
\centering
\includegraphics[width=1\textwidth]{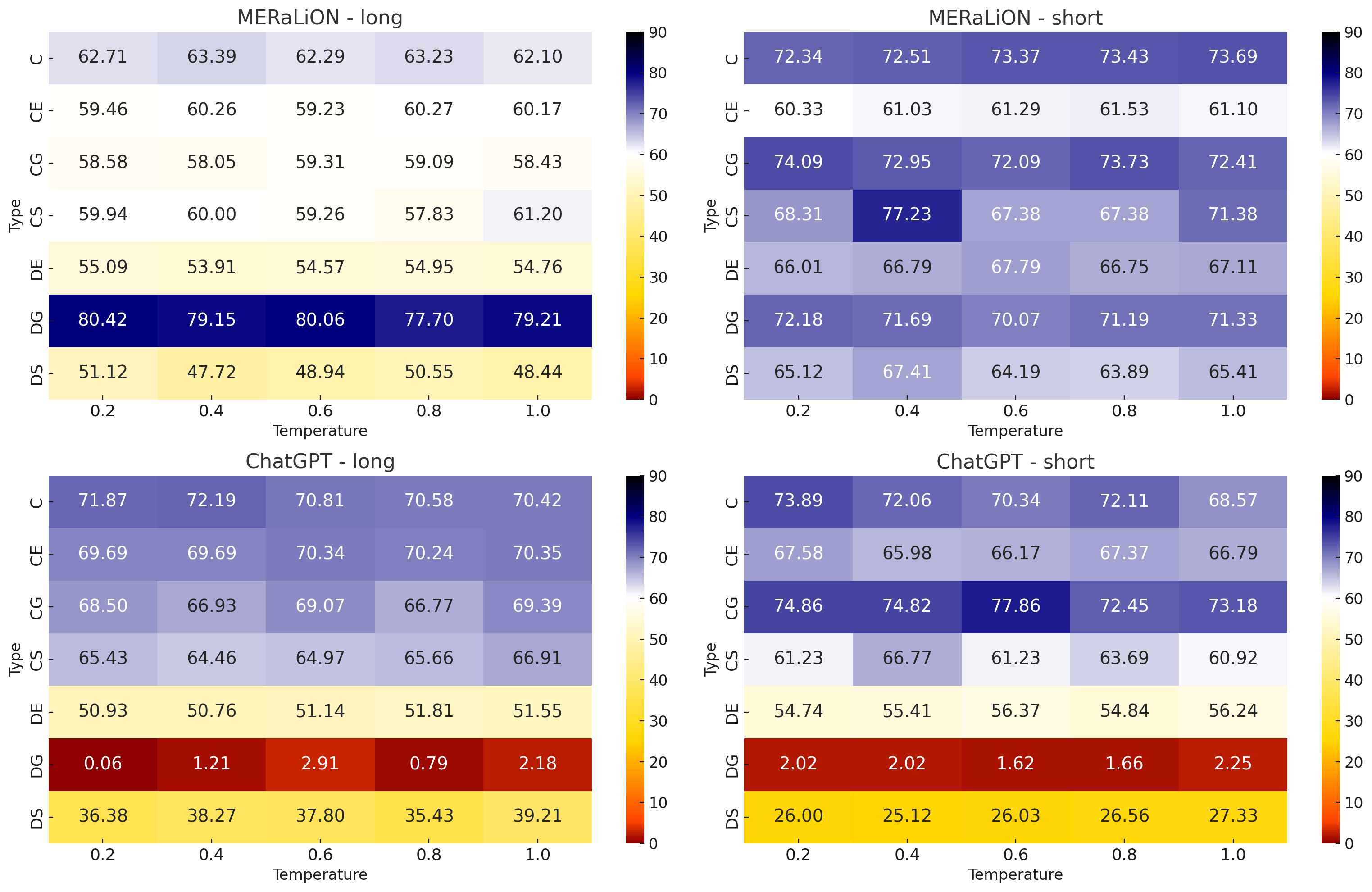}
\caption{Performance of MERaLiON and GPT modes with various temperature settings on the long and short evaluation sets.}
\label{fig:fig_temperature}
\end{figure*}

Qwen achieves the highest accuracy on DP-category questions in the long set but shows a substantial decline in the short set, suggesting a relative strength in directly identifying paralinguistic attributes from long speech. In contrast, Qwen performs noticeably worse than other speech-LLM models on C and CP-category questions.
The two cascaded models generally underperform compared to the multimodal speech-LLMs on both the CP and DP-category questions. This is likely due to their reliance on text-only input, which limits their access to paralinguistic cues present in the original audio. However, their performance shows inconsistencies between the long and short sets, along with notable variation across different question categories. We reserve a definitive interpretation for future analysis on fine-grained types.

Overall, MERaLiON and Gemini demonstrate the most balanced performance across all categories. Gemini achieves the highest  overall performance on the long set, while MERaLion performs the best on the short set. 

\subsubsection{Investigation across Models on Fine-Grained Question Types}
We conduct a more detailed analysis of individual question types within DP and CP categories to gain deeper insights, as illustrated in Figure~\ref{fig:all_model}. There is significant performance divergence across question types in the DP category, showing varying strengths and weaknesses in detecting emotion, gender, and speaker count. Notably, performance disparities among models tend to reduce when contextual information is incorporated, as observed in the CE, CG, and CS question types.

In the CP question category, GPT consistently achieves superior performance on both CE and CG question types across the long and short sets, reflecting its robust capability to reason about emotion and gender within speech contexts. However, MERaLiON notably surpasses GPT on CS-type questions in the short set, underscoring its particular strength in handling speaker-related contextual reasoning with shorter speech samples. Both Gemini and MERaLiON demonstrate competitive, mid-range performance across CP-category questions, often closely following GPT and occasionally  surpassing other models. In contrast, Qwen consistently underperforms across all question types in the CP category. CS-type questions exhibit the highest variability, particularly in the short set, with large performance fluctuations between models. This variability is likely due to the small sample size, which limits statistical reliability. Nevertheless, we include the results for completeness and analysis. Notably, Cascade 2 achieves a performance peak on CS-type questions in the short set. The small number of instances also contributes to an inflated CP score for Cascade 2 in Table~\ref{tab:models_comparison_group}. In the long set, MERaLiON and Gemini perform comparably. Cascade 2's surprisingly strong performance on CE-type questions indicates the limitations of current speech-LLM models in fully leveraging audio cues, suggesting potential for improvement through enhanced multimodal integration.

The breakdown within DP-category questions reveals significant performance variation across question types. MERaLiON consistently outperforms Gemini on DE and DG-type questions and even surpasses GPT on DE, showing strong and stable direct paralinguistic recognition. Qwen performs notably well on DG-type questions in the long set but sharply declines on short speech inputs, indicating a sensitivity to input length. Both cascaded models perform moderately well on DE-type questions, achieving the highest score for DE in the short set. This suggests that emotional cues are partly retrievable from transcripts alone. Nevertheless, they markedly underperform on DG and DS questions due to their lack of direct access to audio-based speaker information. In fact, all models struggle with DS-type questions, though MERaLiON leads in the short, and Gemini performs the best in the long set.

Across all question types in the CP category, as well as DE and DS types, all models consistently perform better in the short set compared to the long set, with the exception of DG-type questions, where this trend does not hold. This suggests that shorter speech inputs are generally easier for models to handle, possibly due to constraints in attention span, context window limitations, or a reduced need for long-range reasoning.

\subsubsection{Temperature Investigations}

Figure~\ref{fig:fig_temperature} presents a detailed analysis of how LLM decoding temperature impacts performance using GPT and MERaLiON, representing closed- and  open-source models, respectively.  GPT was selected over Gemini as the closed-source model due to its stronger performance on CP question categories, which are central to this study. The results show that temperature tuning significantly influences model accuracy in certain categories, revealing distinct patterns within each model and question type.

For C-type questions, GPT performs better at lower temperatures when processing long-form content, whereas MERaLiON benefits more from higher temperatures in the evaluation set with shorter speech. Temperature has minimal impact on CE-type questions for both models, suggesting that reasoning involving emotion and content is less sensitive to response diversity. GPT's performance on CG-type questions is more temperature-dependent, peaking at mid-range values, while MERaLiON remains consistently favors lower settings.

MERaLiON consistently outperforms GPT on DE-type questions, with both models showing moderate improvement at higher temperatures. In DG-type questions, GPT struggles significantly across all temperatures due to content filtering restrictions, whereas MERaLiON maintains top performance with only slight degradation at elevated temperatures. For CS-type questions, MERaLiON experiences the largest performance gains with increasing temperature on the short set. GPT, however, displays a different behavior, with a peak at temperature 0.4 followed by a sharp decline. Our analysis reveals distinct optimal temperature ranges for MERaLiON and GPT depending on input length and question type. 

\section{Summary and Conclusion}
 We propose CP-Bench, a benchmark for evaluating speech-LLMs on contextual paralinguistic reasoning---the integration of verbal content with non-verbal cues like emotion and prosody. The benchmark includes two curated QA datasets that require both linguistic and empathetic understanding. CP-Bench challenges speech-LLMs to demonstrate linguistic comprehension, speaker information, and empathetic insight. Our comprehensive evaluation of leading speech-LLMs, including an analysis of temperature tuning effects, reveals critical strengths and limitations. This work fills an important gap in current benchmarks and provides valuable guidance towards developing more contextually aware and emotionally perceptive speech-based language models.
 
For future work, we plan to expand CP-Bench by adding more diverse evaluation sets, including different languages, and by increasing the density of QA pairs for each question type, especially within the multi-speaker subset, to strengthen the statistical reliability of results. We will also explore improved evaluation techniques for such multimodal reasoning, including methods to enhance the robustness and interpretability of LLM-based judgments. Finally, to further ensure reliability, we  plan to incorporate human reviews to calibrate and validate LLM-based evaluations.

\section*{Limitations}
This study has a few limitations. First, most speech samples contain a single type of emotion, which limits the evaluation of models' ability to handle mixed-emotion inputs. Second, our human annotators generated QA pairs independently resulting in no overlap. This may introduce a degree of bias related to how emotion labels were annotated. However, this lack of overlap does not affect most other QA pairs, which benefit from greater diversity in question types. Third, the number of CS-type questions in the short evaluation set is relatively small, making conclusions drawn for this category less reliable. Finally, our analysis relies on a single LLM-based judge, which may introduce uncertainty due to potential biases or variability in model outputs. Incorporating multiple LLM judges or human evaluators in future work could enhance the robustness and reliability of the evaluation.
\section*{Ethics Statement}
\label{sec: ethics statement}
The speech data used in this study was sourced from publicly available YouTube videos. The speech samples are solely for research and evaluation purposes. Only the URLs linking to the original YouTube videos that correspond to the speech samples will be released. This approach respects content ownership and aligns with platform policies regarding the use of its content.

\section*{Acknowledgments}
This research/project is supported by the National Research Foundation, Singapore under its National Large Language Models Funding Initiative. Any opinions, findings, conclusions, or recommendations expressed in this material are those of the author(s) and do not reflect the views of the National Research Foundation, Singapore.

The authors would also like to thank Yingxu He, Xunlong Zou, Siti Maryam Binte Ahmad Subaidi, Siti Umairah Md Salleh, and Xi Wang for their valuable assistance in this work.

\bibliography{custom}

\appendix
\section{Appendix}
\subsection{Ablation Studies}
\subsubsection{
The use of Paralinguistic Cues in Reasoning for CP questions
}
\label{sssec:swap_emotion}
To investigate whether speech-LLMs reason over paralinguistic cues for CP questions or rely solely on content, we conducted a targeted experiment on CE-type questions. We sampled 27 QA pairs from the short set that explicitly reference emotions. For each pair, we manually constructed a conflict test by altering the textual query to contradict the emotion conveyed in the audio (e.g., replacing a negative emotion term with ``happy'' and a positive emotion term with ``sad''; see Table \ref{tab:ce30_qa_mod}). GPT was then used to evaluate these conflict cases.
\begin{table*}[!t]
\centering
\resizebox{\textwidth}{!}{%
\begin{tabular}{lll}
\toprule
\textbf{Original question} & \textbf{True emotion} & \textbf{Modified emotion} \\
\midrule
Why is the first speaker \st{angry} \textbf{happy}?
& angry & happy \\
Why is the second female speaker talking in a \st{low} \textbf{happy} tone?
&fear	&happy \\
Why does the speaker sound \st{frustrated} \textbf{happy}?
&frustrated  &happy \\
What is the speaker expressing his \st{disgust} \textbf{happiness} about?
&disgusted	&happy \\
Why are the speakers \st{happy} \textbf{sad}?
&happy	&sad \\
Why is the speaker \st{excited} \textbf{sad}?
&excited	&sad \\

\bottomrule
\end{tabular}
}
\caption{Examples of emotion-conflicting samples.}
\label{tab:ce30_qa_mod}
\end{table*}

The model resisted the conflicting query in 17 of 27 cases, correctly grounding its answer in the audio’s emotional cues. In the remaining 10 cases, the model produced inconsistent or hallucinated responses, reflecting difficulty in resolving cross-modal conflicts. This experiment demonstrates that for a significant portion of CE questions, the model can often leverage paralinguistic evidence beyond mere content and incorporate cues from the audio.

\subsubsection{Statistical Reliability of the LLaMa Judge}

To assess the stability of the LLaMa judge, we run it five times with the default temperature setting for every speech-LLM response. We then average the scores and analyze the standard deviations (std) of the results. We also compare these with scores from a single run under greedy decoding (temperature$=0$). Finally, we conduct a human evaluation experiment on a small subset for further validation.

\begin{table}[t]
\centering
\caption{Scores of MERaLiON-generated answers evaluated with the LLaMA judge. Results are reported as the mean $\pm$ std over five runs with the default temperature, and from a single run with greedy decoding (GD; temperature = 0).}

\begin{subtable}[t]{0.48\textwidth} 
        \centering
        \caption{Long set}
        \label{tab:std_long}
\begin{tabular}{c|r|r}
\hline
Type & Default temp. & GD \\
\hline
C & 62.10$\pm$ 1.73& 61.13\\
CE& 60.17$\pm$  0.44& 60.59\\
CG& 58.43$\pm$  1.96& 58.74\\
CS& 61.20$\pm$  1.77& 62.57\\
DE& 54.76$\pm$  0.57& 55.74\\
DG& 79.21$\pm$  0.55& 76.97\\
DS& 48.44$\pm$  0.51& 48.50\\
\hline
\end{tabular}
\end{subtable}
\hfill 

\begin{subtable}[t]{0.48\textwidth} 
        \centering
        \caption{Short set}
        \label{std_short}
\begin{tabular}{c|r|r}
\hline
Type & Default temp. & GD \\
\hline
C & 73.69$\pm$ 0.97& 74.86\\
CE& 61.10$\pm$ 0.16& 61.19\\
CG& 72.41$\pm$ 1.11& 73.18\\
CS& 71.38$\pm$ 2.57& 69.23\\
DE& 67.11$\pm$ 0.52& 66.88\\
DG& 71.33$\pm$ 0.47& 71.91\\
DS& 65.41$\pm$ 0.35& 65.33\\

\hline
\end{tabular}
    \end{subtable}

\label{tab:std}
\end{table}

As shown in Table \ref{tab:std}, score variance across runs is minimal, with std values of 0.44--1.73 and 0.35--2.57 for the seven question types answered by MERaLiON in the long and short sets, respectively. Moreover, average performance under default settings is highly consistent with the greedy decoding results, showing an average absolute difference of only 0.91 (long set) and 0.72 (short set).


For the human evaluation, we used the same 27 CP questions from the short set (see \ref{sssec:swap_emotion}) with MERaLiON’s generated answers. Three human annotators independently rated each answer following the same scoring guidelines as the LLM judge (Figure~\ref{fig:prompt_qa_dim}).   The averaged human scores and the LLaMa judge scores averaged over five runs are positively correlated, with a coefficient of 0.71, indicating a reasonable alignment between prompt-based LLM judgments and human evaluations.

These findings---low variance across runs, strong agreement with greedy decoding, and the consistency with human ratings---demonstrate that our prompt-based evaluation is stable and well-suited for large-scale speech-LLM benchmarking.

\subsubsection{Reliance on Content Cues in DG Question Reasoning }
Cascaded model 2 showed good performance on DG in the long set (see Figure~\ref{fig:all_model}. To better understand this result, we conducted an error analysis of the LLM judge’s reasoning on a subset of 33 high-scoring DG pairs produced by Cascade Model 2.

Our analysis indicates the LLM primarily relies on explicit textual cues: (1) gender-specific pronouns, titles, nouns, and names, and (2) stereotypical associations and phrasing styles. Among the 33 answers, 6 were based on cue type (1), 7 on cue type (2), and 5 on both. In 15 cases, the model failed to infer gender; 5 of these were polar questions answered correctly by coincidence (e.g., “no” when asked if a specific gender was present).

These findings suggest that the cascaded model’s success on DG questions largely reflects shallow content cues rather than genuine speaker inference. Such cues may yield correct answers at times, but gender is independent of mentioned genders, and stereotypes are unreliable. Hence, content-based evidence alone is not a robust basis for gender identification.

\begin{figure*}[t]
    \raggedright 
    {\small
    \begin{tcolorbox}[colframe=black, colback=gray!10, arc=2mm, boxrule=0.5pt, width=1\textwidth]
    Q: How would you describe the tone of the speaker's voice?\\
A: The speaker's tone sounds \#XXXX\#.\\
\\
Q: What emotions do you think the speaker is expressing?\\
A: Based on the tone of the speaker's voice and their delivery, it seems like the speaker is \#XXXX\#.\\
\\
Q: Do you think the speaker sounds happy, sad, angry, or something else?\\
A: The speaker's emotion seems \#XXXX\#.\\
\\
Q: Can you describe the tone of the speaker's voice and what it suggests about their emotion?\\
A: The tone of the speaker's voice suggests that the speaker may be feeling \#XXXX\#.\\
\\
Q: Based on what you heard, what emotion do you think the speaker is conveying?\\
A: Based on the auditory cues, it sounds like the speaker is \#XXXX\#.\\
\\
Q: What do you think the speaker is feeling based on their speech patterns (e.g., pace, pitch, volume)?\\
A: Based on the speaker's speech patterns, it seems like they are feeling \#XXXX\#.\\
\\
Q: How would you interpret the speaker's emotional state from their speech?\\
A: From the speaker's speech, it seems they are in a \#XXXX\# state.\\
\\
Q: Based on the speaker's speech patterns, what do you think they are feeling?\\
A: The speaker's speech suggests they might be feeling \#XXXX\#.\\

   \end{tcolorbox}

    }\vspace{-2pt}
   \caption{The QA template for DE-type QA generation.}
    \label{fig:de_qa}
\end{figure*}

\subsection{DE-type QA generation template}
\label{ssec:appendix_prompts}
For DE-type QA generation, we generate one question per speech sample, querying the emotion conveyed in the speech, using a QA template randomly selected from those shown in Figure~\ref{fig:de_qa}. ``\#XXXX\#'' is replaced by the emotion labels of the speech samples.

\subsection{Prompt templates}
\subsubsection{Generation}
\label{sssec:ans_prompt}
\begin{figure*}[h]
    \raggedright
    {\small
    \begin{tcolorbox}[colframe=black, colback=gray!10, arc=2mm, boxrule=0.5pt, width=1\textwidth, sharp corners=south]

\begin{verbatim}
chat_prompt = [ 
    { 
        "role": "user", 
        "content": [ 
            { 
                "type": "text", 
                "text": question 
            }, 
            { 
                "type": "input_audio", 
                "input_audio": { 
                    "data": encoded_string, 
                    "format": "wav" 
                } 
            } 
        ] 
    } 
]
\end{verbatim}
    \end{tcolorbox}
    }
    \vspace{-2pt}
    \caption{Example prompt structure including textual instruction and embedded audio data for multimodal input.}
    \label{fig:prompt_audio_input}
\end{figure*}

Prompt to generate answer from speech-LLM is shown in Figure~\ref{fig:prompt_audio_input}.

\subsubsection{Judge}
\label{ssec:appendix_prompts}
We adapted the judging prompt from AudioBench~\cite{audiobench} to better suit contextual-paralinguistic question answering, with tailored modifications to the scoring criteria. In particular, we refined the definitions to emphasize paralinguistic reasoning and reward responses that are both accurate and concise. The revised scoring rubric is presented in Figure~\ref{fig:prompt_qa_dim}.
\begin{figure*}[h]
    \raggedright  
    {\small
    \begin{tcolorbox}[colframe=black, colback=gray!10, arc=2mm, boxrule=0.5pt, width=1\textwidth]
Rate the model's answer based on its alignment with the reference answer, considering both speech content and paralinguistic information, including emotion, speaker identity, speaker transitions, number of speakers, gender cues, tone, intent, and overall feeling. Focus on both what is said and how it is conveyed. Be critical of details, and if the model response is vague or says something like "cannot decide," please rate it as 0.
Evaluation Criteria:Assess whether the model's response mirrors the reference in terms of content accuracy, paralinguistic relevance, and alignment with speech-specific nuances such as emotion, intent, and speaker characteristics.
\begin{itemize}
    \item Score 0: The answer refuses to provide concrete results, saying ``cannot decide'' or similar.
    \item Score 0: The answer is completely misaligned, providing incorrect, misleading, or irrelevant information regarding the speech content or paralinguistic elements.
    \item Score 1: The answer minimally aligns with the reference, often misunderstanding the speech or missing important paralinguistic cues (e.g., misidentifying speaker emotion or gender).
    \item Score 2: The answer recognizes the general topic but diverges significantly from the reference in accuracy, relevance, or paralinguistic understanding.
    \item Score 3: The answer is generally aligned with the reference but lacks detail or precise accuracy in either speech content or paralinguistic aspects or both together.
    \item Score 4: The answer is mostly accurate and relevant, closely following the reference but could be clearer or more detailed in either content or paralinguistic interpretation.
    \item Score 5: The answer is highly accurate, detailed, and matches the reference answer perfectly, capturing either content meaning or paralinguistic essence (emotion, tone, speaker identity, number of speaker, etc.) or both together.
\end{itemize}
    \end{tcolorbox}
    }
    \vspace{-2pt}
   \caption{The scoring definition in the prompt for judging.}
    \label{fig:prompt_qa_dim}
\end{figure*}

\subsection{
    Annotation guidelines
}
\label{ssec:anno_guideline}
All three human annotators followed a unified guideline (see Figure~\ref{fig:guideline_label}) when correcting gender and emotion labels and cross-discussed ambiguous cases. This protocol helped ensure consistency and reduced individual bias even though no separate verification round was conducted after correction. QA generation guideline for human annotator is shown in Figure~\ref{fig:guideline_qa}.

\begin{figure*}[h]
    \raggedright 
    {\small
    \begin{tcolorbox}[colframe=black, colback=gray!10, arc=2mm, boxrule=0.5pt, width=1\textwidth]

\underline{\textbf{Task: label correction}}\\
\textbf{Objective:} Correct the pre-assigned emotion and gender labels.\\
\textbf{Input:} Two JSON files are provided, containing:
    \begin{itemize}
    \item Emotion Labels: Annotated for every 2-second segment in speech samples.
    \item Gender Labels: Annotated for every 2-second segment in speech samples.
    \end{itemize}
\textbf{Instructions:}
    \begin{itemize}
    \item Listen to the corresponding segments in the speech samples and validate or correct the provided labels.
    \item You may use the transcript for additional content understanding.
    \item Ensure consistency and accuracy when correcting labels.
    \item Consider the context of the conversation or narrative to avoid mislabeling emotions. For example, a person may raise their voice due to excitement (happy) rather than anger.
    \item If a segment expresses mixed emotions, prioritize the dominant emotion.
    \item If a label cannot be confidently corrected, make a note of the ambiguity for review.
    \item You can use any tools of your choice as you will need to select durations based on JSON file to correct/check the emotion and gender labels. Open-source tools like Audacity, WaveSurfer can also work well. 
    \end{itemize}
\textbf{Outcome:} JSON files with corrected emotion and gender labels.\\

   \end{tcolorbox}

    }
    \vspace{-2pt}
   \caption{Annotation guideline - Label correction.}
    \label{fig:guideline_label}
\end{figure*}

\begin{figure*}[h]
    \raggedright 
    {\small
    \begin{tcolorbox}[colframe=black, colback=gray!10, arc=2mm, boxrule=0.5pt, width=1\textwidth]
\underline{\textbf{Task: Question Generation}}\\
\textbf{Objective:} Create paralinguistic-based questions and answers for each speech sample. These questions should involve reasoning and cover both paralinguistic (emotion, gender, sentiment, relationships) and content information (what is spoken)\\
\textbf{Instructions:}
    \begin{enumerate}
    \item Listen to the entire speech sample to understand its context and nuances.
    \item You may use the transcript for additional content understanding, but ensure questions focus on both paralinguistic and content information.
    \item Create at least two questions, with at least one being a CE, CE+, or CG-type question.
    \item Label each question with its question type (from Appendix B).
    \item 	Use diverse question types for variety. 
    \item 	Refer to corrected labels (emotion, gender) from Task 1 for guidance.
    \item 	Question generation requires listening to the entire speech file (and pausing when necessary). You can use the knowledge of emotion and gender obtained from the label correction task.
    \item 	Focus on the clip context only: 
    \begin{itemize}
        \item 	Most clips are extracted from YouTube videos featuring drama, conversations, experience sharing, general discussions, etc.
        \item 	You are not required to consider or analyze the context of the entire YouTube video.
        \item 	Questions should be generated solely based on the specific clip provided, ensuring the content and paralinguistics are derived from that segment alone
    \end{itemize}
    \item 	Handling clips with factual information:
    \begin{itemize}
        \item 	If the clip primarily contains factual information (e.g., a conversation discussing Singapore politics or explaining historical events), you may use background knowledge to generate appropriate questions.
        \item 	However, take special care to ensure such questions remain aligned with the core objective of focusing on paralinguistics-related spoken QA (e.g., emotion, tone, relationships, or sentiment inferred from the speech).
        \item 	Make a note of such cases where factual information dominates, so they can be reviewed separately to ensure alignment with the project’s goals. Avoid steering away from the paralinguistics emphasis.
    \end{itemize}
    \item 	If a clip does not have enough content, discard the audio. 
    \item 	Background knowledge should only be used minimally to supplement understanding in factual discussions, not as the primary basis for generating questions. The focus should remain on cues within the clip.
    \item 	Please ensure the English in the QA text is correct. You may use ChatGPT to help with language correction.
    \end{enumerate}
\textbf{Outcome:} question-answer pairs and question type for each speech samples.

   \end{tcolorbox}

    }
    \vspace{-2pt}
   \caption{Annotation guideline - QA generation.}
    \label{fig:guideline_qa}
\end{figure*}

\subsection{Model prediction examples}
Table \ref{tab:qa_examples} and \ref{tab:qa_examples_meralion} show samples of generated answers from GPT and MERaLiON models using temperature 0.7 and 1.0, respectively (default setting). It is interesting to note that both models provide correct answers, especially  for contextual types of questions. 
GPT tends to provide more elaborate and user-satisfying answers. MERaLiON answers are generally more concise and direct. GPT notably exhibits a common failure to answer privacy-related questions due to its limitations. 
These observations offer initial insights into model strengths and weaknesses, and a more exhaustive qualitative analysis will be a key focus of our future work.

\begin{table*}[t]
\centering
\small
\begin{tabular}{|p{0.20\linewidth}|p{0.35\linewidth}|p{0.35\linewidth}|}
\hline
\textbf{Question} & \textbf{Human Answer} & \textbf{GPT Answer} \\
\hline
How does the female speaker feel about choosing the chicken rice spot? &
She feels happy and confident in her choice because she frequently visited the shop in Toa Payoh as it is near her area, and it is good even though it is not super well known. &
The female speaker seems excited and confident about choosing her favorite chicken rice spot. She mentions it's in Toa Payoh, and although it may not be well-known, she believes it's very good. Her enthusiasm suggests she has a positive attitude toward her choice. \\
\hline
How did the man describe the pilot's voice and what was the problem that the pilot reported? &
The man described the pilot as sounding "very scared" when he was reporting about the flight control problem that he probably was facing. &
The pilot's voice was described as "very scared" when he reported the problem. The issue reported by the pilot was a flight control problem. The pilot had initially requested to increase altitude and then mentioned the control issue before starting to make a right turn to return to the airport. The plane then vanished off the radar. \\
\hline
Why is the singer, Sezairi happy in this context? &
He is happy as he just performed his original song and gets to share the story behind the song. &
Sezairi is happy in this context because he created a song, "It's You," as a way to express feelings he couldn't convey during his wedding vows. He found a creative outlet in songwriting to share his emotions, which is a joyful and fulfilling experience for him. Additionally, he seems excited about his upcoming music releases and projects, indicating a positive and enthusiastic outlook on his career and personal life. \\
\hline
Why is the speaker angry? &
He is angry because his taxi driver does not know how to go to a certain place and he does not know how to use a GPS. &
The speaker is angry because a taxi driver did not know how to get to a certain place and did not use a GPS to find the way. The speaker feels that it is the driver's job to know how to navigate to destinations or use tools like GPS if they are unsure. The frustration comes from the perceived lack of competence and preparedness on the part of the taxi driver. \\
\hline
What is the relationship between the two speakers in the clip? &
They are grandfather and grandson. &
Based on the content of the clip, it appears that the relationship between the two speakers is that of a grandparent and grandchild. The grandchild addresses the elder as "gong gong," which is a term used in some Chinese dialects to refer to a grandfather. The interaction suggests a familial relationship where the younger person is guiding the elder on how to prevent dengue fever, indicating a dynamic where the grandchild is taking on a protective or educational role. \\
\hline
What emotions prompt the two speakers to sing outside of mass? &
The woman is prompted to sing outside mass when she feels happy and thankful to God but the man is prompted to sing when he is sad or fearful. &
The two speakers are prompted to sing outside of mass due to various emotions. Auntie Janice sings when she is feeling grateful and wants to thank God for a wonderful day. Uncle Mark, on the other hand, sings when he is feeling down or scared, as a way to remind himself that God is his strength and to feel better. \\
\hline
\end{tabular}
\caption{Examples of contextual-paralinguistic questions with human and GPT answers.}
\label{tab:qa_examples}
\end{table*}

\begin{table*}[t]
\centering
\small
\begin{tabular}{|p{0.20\linewidth}|p{0.35\linewidth}|p{0.35\linewidth}|}
\hline
\textbf{Question} & \textbf{Human Answer} & \textbf{MERaLiON Answer} \\
\hline
Why did the speaker feel sad? & 
The speaker felt sad as the number of deaths in care homes were significantly high as more patients passed away. & 
The speaker felt sad because the number of deaths in care homes was significantly high. \\
\hline
What is the man angry about? & 
He was ranting about a comedian who made racist jokes that could cause tension between Singapore and Malaysia. & 
The man is angry about a joke made by someone named Joscelyn Chia, who he believes is not funny and is trying to cause trouble between Singapore and Malaysia. \\
\hline
What is the relationship between Intan and Karen? & 
Karen is a special guest on Intan's show and also a long time friend of hers. & 
Intan and Karen are long-time friends. \\
\hline
Why is the man happy when asked about how his 2018 went? & 
He was happy as 2018 was the best year for all of them, they finished all of their raya collection launches, their raya sales went great, and the response has been great for their latest collection. & 
The man is happy because 2018 has been a successful year for his company, Jovian Mandagie. They finished all their Raya collection launches and sales went well, especially for their last collection called "Jovian Lux Shawl". He is satisfied with the results and hopes 2019 will be even better. \\
\hline
How does Margaret feel about Lucy's shopping trip? & 
Margaret feels surprised and concerned that Lucy is going out when she should still be avoiding crowded places. & 
Margaret is concerned about Lucy shopping in crowded places, especially since the face mask rules have been relaxed. She advises Lucy to avoid crowded areas and be careful. \\
\hline
What emotions prompt the two speakers to sing outside of mass? & 
The woman is prompted to sing outside mass when she feels happy and thankful to God but the man is prompted to sing when he is sad or fearful. & 
The two speakers sing outside of mass when they are feeling: * **Untie Janice:** Grateful and wanting to thank God for a wonderful day. 
 * **Uncle Mark:** Down or scared, seeking comfort and strength in God. \\
\hline
\end{tabular}
\caption{Examples of contextual-paralinguistic questions with human and MERaLiON answers.}
\label{tab:qa_examples_meralion}
\end{table*}

\end{document}